\documentclass[10pt,journal,compsoc]{IEEEtran}
\usepackage[nocompress]{cite}
\usepackage[pdftex]{graphicx}
\usepackage{amsmath}
\usepackage{algorithmic}
\usepackage{array}
\usepackage[caption=false,font=footnotesize,labelfont=sf,textfont=sf]{subfig}
\usepackage{url}
\usepackage[colorlinks=true]{hyperref}
\usepackage{ragged2e}
\usepackage{amssymb}

\begin{document}

\title{MMC: Multi-Modal Colorization of Images using Textual Descriptions}
\author{
  Subhankar~Ghosh,
  Saumik~Bhattacharya,
  Prasun~Roy,
  Umapada~Pal,
  and~Michael~Blumenstein
  \IEEEcompsocitemizethanks{
     \IEEEcompsocthanksitem  S. Ghosh P. Roy, and M. Blumenstein are with Faculty of Engineering and IT,University of Technology Sydney, NSW, Australia
    \IEEEcompsocthanksitem  U. Pal is with the Computer Vision and Pattern Recognition Unit, Indian Statistical Institute, Kolkata, India.
    \IEEEcompsocthanksitem S. Bhattacharya is with Indian Institute of Technology, Kharagpur, India.
  }
}

\IEEEtitleabstractindextext{
\begin{abstract}
 \justifying{ Handling various objects with different colors is a significant challenge for image colorization techniques. Thus, for complex real-world scenes, the existing image colorization algorithms often fail to maintain color consistency. In this work, we attempt to integrate textual descriptions as an auxiliary condition, along with the grayscale image that is to be colorized, to improve the fidelity of the colorization process. To do so, we have proposed a  deep network that takes two inputs (grayscale image and the respective encoded text description) and tries to predict the relevant color components. Also, we have predicted each object in the image and have colorized them with their individual description to incorporate their specific attributes in the colorization process. After that, a fusion model fuses all the image objects (segments) to generate the final colorized image. As the respective textual descriptions contain color information of the objects present in the image, text encoding helps to improve the overall quality of predicted colors. In terms of performance, the proposed method outperforms existing colorization techniques in terms of LPIPS, PSNR and SSIM metrics.}
\end{abstract}

}

\maketitle

\section{Introduction}
\label{sec:intro}

\begin{figure}
  \centering
  \includegraphics[width=\linewidth ]{ 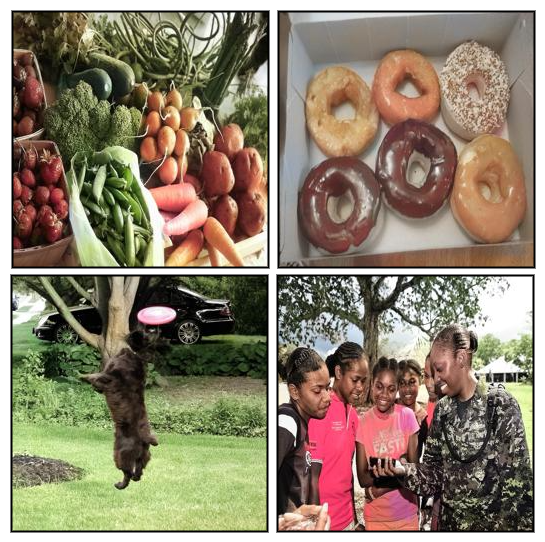}
  \caption{Examples of some colorized outputs from our multimodal colorization approach. [Best visible in $300\%$ zoom ]}
    \label{fig:front_result}
\end{figure}

Colorization injects life into a grayscale image. Usually, in a real-world scene,  different objects contain different colors. Moreover, in certain cases, one object may have different color variations which makes the colorization task complex. By nature, colorization is an ill-posed problem due to this one-to-many association during the colorization process. A variety of colorization techniques has been proposed in recent years, and state-of-the-art performance has been reported on current databases even for complex real-world scenes \cite{Caesar2018COCOStuffTA,WelinderEtal2010,Deng2009ImageNetAL,Kumar2021ColorizationT,Wu2021TowardsVA,Su2020InstanceAwareIC}. These colorization techniques differ in many aspects, such as network architecture, loss functions, learning strategies, etc. However, the existing colorization processes \cite{Wu2022FinegrainedSE,Huang2022DeepLF,Xiao2022SemanticawareAI,Luo2022ThermalII,Treneska2022GANBasedIC} mostly follow unconditional generation where the colors are predicted only from the input grayscale image. In this work, we propose a colorization network that colorizes the image as shown in Fig.\ref{fig:front_result}, using multi-modal feature attention. In the proposed method, we subdivide the task into two parts to colorize the entire image. At first, we find the mask of each object in the image to associate each object in a scene with an instance. In our multimodal colorization approach, we consider a textual description of these instances along with the grayscale image in the colorization process. The detected object is colorized by the instance object colorization (IOC) module conditioned over the grayscale image and the corresponding language description. To achieve a superior output, we design the IOC module as a multi-task network that predicts the class of the object instances along with its colorization to make sure that the IOC module closely learns the association between an object and its color. As we are using text information as an auxiliary condition in the process, the ambiguity in the object-color association is greatly reduced. After the instance-level colorization, we combine all the objects into an image considering their previous spatial support. Finally, we train a network that takes the partially colorized image generated from the previous stage as input, and the corresponding language description of the entire image (including background and non-object instances like `sky', `field' etc.) of the image to generate the final colorized image.\\
Our contributions are as follows.\\
- The proposed IOC module utilizes instance label image colorization exploiting object-color associations. To achieve superior performance, we design the IOC module as a multi-task network. To the best of our knowledge, this is the first attempt to design a multi-task network for the colorization task, considering the object-level instances.\\  
- A multi-modal pipeline is proposed that colorizes the image using language information, which is considered as auxiliary conditioning in the colorization process.\\
- To ensure high fidelity over colors, a novel loss function is proposed that captures the overall colorfulness of a scene.\\\\
The rest of the paper is organized as follows. In Sec. \ref{sec:relworks}, we discuss existing work related to the image colorization task. The overview of the pipeline and the proposed methodology are discussed in Sec. \ref{sec:overview} and Sec. \ref{sec:method}, respectively. Sec. \ref{sec:results}, discusses different experimental results including dataset, qualitative result, comparative results with existing methods, different ablation studies etc. Finally, we conclude the paper in Sec. \ref{sec:conclusion} discussing the major observations, limitations, and future scope of the proposed algorithm.

\section{Related work}
\label{sec:relworks}
In the last two decades, image colorizations have been  one of the main focus areas of computer vision research.Most of these methods were influenced by conventional machine learning approaches \cite{Levin2004ColorizationUO,Huang2005AnAE,Bastos2013RUNTIMEGS}.  In the last few years, the trend has shifted to deep learning (DL)-based approaches due to the success of those approaches in different fields \cite{anwar2020ColorSurvey,Wang2018GeneratingHQ,Tola2008AFL,Perazzi2016ABD}.  Recently, DL-enabled automatic image colorization systems  have  shown  an  impressive  performance  for the colorization task \cite{Zhang2016ColorfulIC,Carlucci2018DE2CODD,Levin2004ColorizationUO,Cheng2015DeepC,Wang2018GeneratingHQ,Bahng2018ColoringWW,Su2020InstanceAwareIC,Kumar2021ColorizationT}. 

Deep colorization \cite{Cheng2015DeepC} was the first image colorization method using Deep learning. In the architecture, the authors design a network that contains five fully connected layers with ReLU activation. In training, the least-squares error was used as a loss function. \\
Deep depth colorization \cite{Carlucci2018DE2CODD} used the deep depth information of the image, which is generated from the pre-trained Image-Net\cite{Deng2009ImageNetAL} networks. Pre-trained  weights  are  kept  frozen  in  the  network,  and this  pre-trained  network  is  merely  used  as  a  feature  extractor.

The text2color \cite{Bahng2018ColoringWW} model consists of two conditional adversarial networks: the Text to Palette Generation network and the Palette-based colorization network. In order to train the Text to Palette Generation network, the palette dataset is used as well as the text dataset. Text to Palette Generation networks identify the fake and real color palettes based on the text. This network uses Huber loss as a loss function. U-NET architecture is used to design a palette-based colorization network in which a color palette is used as a conditional input. In order to classify the colorized image as real or fake, the authors used a series of convo2d and LeakyRelu \cite{Maas2013RectifierNI} modules. In this method, the number of color palettes is 4 to 6. The generated color image entirely depends on the number of the color palette.  \\
According to Zhang et al. \cite{Zhang2016ColorfulIC}, the authors proposed the first CNN architecture for colorization of the images. Using a cross-channel encoder, the authors colorized their data. Objects cannot be colored with the appropriate color using the colorization method.\\
In the Instance-aware image colorization\cite{Su2020InstanceAwareIC}, the author presented the colorization method that colorized a wide range of multiple objects with diverse contexts. The backbone of the network was taken from Zhang et al. \cite{Zhang2017RealtimeUI}.\\
Towards Vivid and Diverse Image Colorization with Generative Color Prior \cite{Wu2021TowardsVA}, the authors first used a pre-trained GAN for the feature matching. After that the authors generated the various colors by changing the latent space for the next GAN network. If the pre-trained GAN produced misleading features, then the colorization network generated the unnatural colorization of the image. \\
In Colorization transfer\cite{Kumar2021ColorizationT}, the author used conditional an auto regressive transformer for generations the image in low resolution. After that, the network consists of two parallel networks, one for coarse colorization and one for fine colorization.\\
In L-code \cite{Weng2022LCoDeLC} language-based colorization network is proposed using the color corresponding matrix. Also, the authors used the soft-gated injection module.
Though there are several algorithms available for image colorization, to the best of our knowledge, none of them exploit a multi-modal approach at the instance-level for the colorization task. 

\begin{figure}
  \centering
  \includegraphics[width=\linewidth ]{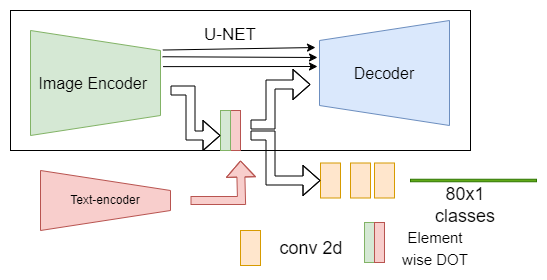}
  \caption{Block diagram of the instance object colorization (IOC) module.}
    \label{fig:obj_network}
\end{figure}

\begin{figure*}
  \centering
  \includegraphics[width=\linewidth ]{ 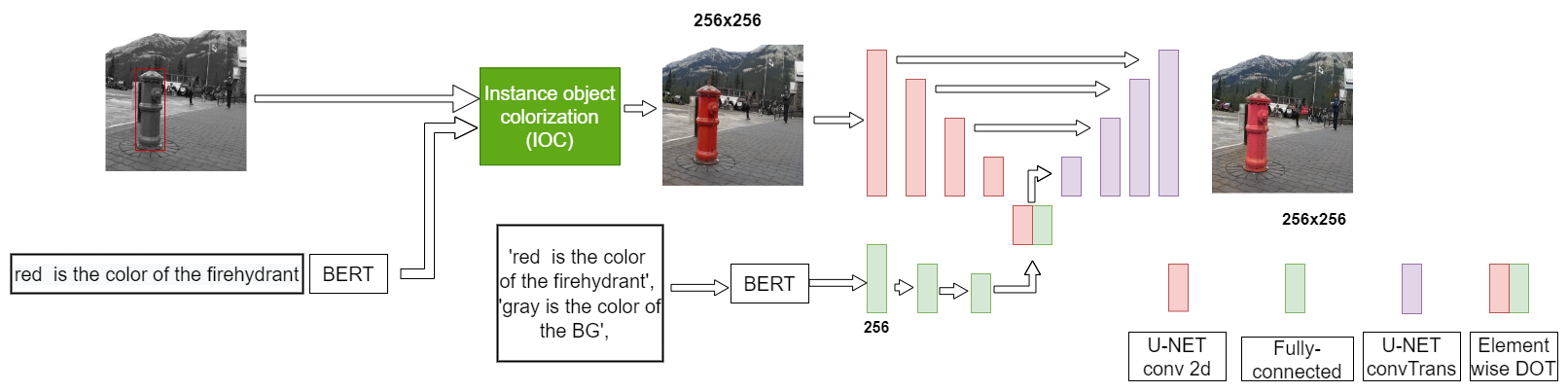}
  \caption{A block diagram of the proposed algorithm. The IOC module colorizes all the object instances, whereas the fusion module takes the partially colored image and generates the fully colorized image. [Best visible in $300\%$ zoom ] }
    \label{fig:fusion_network}
\end{figure*}

\section{Overview}
\label{sec:overview}
The proposed method takes a grayscale image and an image color description as inputs to the network. The textual description contains both object-level descriptions (e.g. 'red ball', 'pink flower' etc.) and background or non-object descriptions (e.g. `blue sky', `green field' etc.). The network predicts 2 missing color channels in the CIE Lab color space. The proposed method has 2 substages. The instance object colorization (IOC) module colorizes object-level instances by solving colorization and classification tasks simultaneously, considering relevant object-level textual descriptions only. In the next stage,  the entire textual color information is passed through the second network along with the partially colorized image to obtain the fully colorized image.

\section{Method}
\label{sec:method}
\subsection{Mask-based Object detection}
For object detection, our method uses the Masked R-CNN \cite{Girdhar2018DetectandTrackEP} network for pixel-accurate object instance marking. After detecting the object's bounding box, we resized the object to a 256x256 resolution. The actual coordinate or the \textit{support} of the object is stored for the fusion network. The detected object is resized and split into two parts- one is the grayscale image that contains the L channel, and another one with the color information containing the `ab' channel.
\subsection{Color information encoding}
One of the key contributions of our work is to encode the color information of the object instances to increase the fidelity in the colorization process. For the color information encoding, we use the BERT\cite{Devlin2019BERTPO} model to convert the textual description of an object to a $256\times 1$  dimensional vector $\mathbf{v}_i$. This instance-level encoding is used by our proposed IOC module.\\
In a complex scene, there might be entities that can be detected as an object by the Masked R-CNN because of their non-object nature (e.g. `sky'), or because the classes are simply not included in the Masked R-CNN training (e.g. `tiger'). However, it is often probable to get color descriptions of these entities fairly easily (e.g., 'blue sky', `tiger with yellow and black stripes'). Thus, we encode all the available auxiliary text information associated with scene using BERT to a $256\times 1$ vector $\mathbf{v}_o$ and pass it to the final fusion module.     
\subsection{Instance object colorization (IOC) Module}
The IOC module uses a modified UNet-like structure as the backbone with two inputs. One of the inputs is the $256\times 256\times 1$  dimensional gray scale object image detected by the Masked R-CNNN, which is passed through the image encoder to generate a feature representation of size $8\times 8\times 64$ at the end of the encoder module. The other input to the IOC module is the color encoding $\mathbf{v}_i$ received from the frozen BERT model. The feature representation $\mathbf{v}_i$ is passed through three fully-connected trainable layers of size 256, 1024 and 4096, respectively. The final output of the fully-connected layer is reshaped to size  $8\times 8\times 64$ and multiplied in an element-wise manner with the output of the image encoder to provide the textual conditioning in the generation pipeline. 
The text-conditioned feature vector is used for the multi-tasking approach. In one of the output branches of the IOC, we try to reconstruct the color information (`ab' channel) of the image using the image decoder. In the other branch of the IOC, we use convolutional layers to classify the object instances among the 80 classes available for the Masked R-CNN module. A schematic of the IOC module has been shown in Fig.\ref{fig:obj_network}. 
The loss functions used to train the model are discussed in Sec. \ref{subsec:loss}.
The proposed encoder contains four 2D 
convolutional layers with ReLu activation. Each of the layers uses a stride of 2 and a batch normalization operation at the end of the layer. In the decoder part, in each layer, we use 2D up-convolution followed by the Batch Normalization and ReLu activation function. Like UNet model, we also use the skip connection from the encoder part to the respective decoder with the same resolution. In the label classifier, there are three 2D convolutional layers followed by one dense layer to find the class label prediction. 
cla
\subsection{Fusion module for colorization }
As shown in Fig.\ref{fig:fusion_network},
A fusion module is used to re-colorize the merged image, which is partly colorized by the Instance object colorization module (IOC). First, we marge the colorized object mask from the IOC module. In the merging technique, the height probabilities of the object color features are included in the merged image. By discarding the low probability object features, the overlap problems are solved.
The fusion module uses a modified UNet-like structure as a backbone of two inputs. One input is a merged image and another is the text encoding of the input image. The size of the marge image is 256x256x3 and the text encoding size is 256.
In the image encoder, four convolutions followed by BN and ReLu activation are used to generate a feature representation with size 8x8x64 at the end of the image encoder with down-sampling by 2. In the text encoding pathway three fully-connected layers with sizes of 256,1024,4096 are used to generate the high-dimensional feature representation. After that, we compute the element-wise dot product with the text-based feature vector and image-based feature vector, which gives the multi-modal attention to the networks. The decoder is designed employing a UNet-like decoder architecture. Four up-convolution layers followed by BN and ReLu activation are used to generate the final output with up-samplings by 2.

\subsection{Discriminator }
For superior generation quality, both the IOC module and the Fusion module use two separate discriminators to check the quality of the generated outputs. In both modules, PatchGan-based discriminator is used to judge the generated image. The input size of the discriminator is 256 x256x3. Three convolutional layers, with 64,128 and 256 filters and ReLU activation are used to extract the features of the input image. The grayscale input image ($L^i$) is stacked  with either a target image ($T^i$) or with the estimated image ($E^i$) where  $T^i$ and $E^i$ are the $AB$ channel of the color image. During the training of the discriminator, the  ($L^i$,$T^i$) stack is labeled as real, and the ($L^i$,$E^i$) stack is labeled as fake.

\subsection{Loss }
\label{subsec:loss}
Both the IOC and the fusion modules are trained independently considering the same sets of losses. We consider that $L^i$ as the input image to module $m$ that has auxiliary text conditioning $S^i$. We have an estimated image $E^i$ whose ground truth is $T^i$. For the IOC module, $E^i$ and $T^i$ are the estimated image and the ground truth image of each object instance, respectively; whereas for the fusion module, $E^i$ and $T^i$ indicate the final estimated image and the actual ground truth of the scene. 
To train the generator, the proposed method uses three different types of losses- $L_1$, perceptual and colorfulness loss.

For overall pixel-level fidelity, we consider the $L_1$ loss for the generation process. The $L_1$ loss for the generator is calculated as:
\begin{equation}
\mathcal{L}^G_1=\|E^i-T^i\|_1=\|G_m(L^i,S^i)-T^i\|_1
\end{equation}
where $G_m$ indicates the generator of module $m$ that maps $L^i$ and $S^i$ to $E^i$.\\
To maintain high perceptual quality of the generated image, we introduce a perceptual loss term while training the generator $G_m$. We have calculated the perceptual loss $\mathcal{L}_{per}^{G_m}$ as
\begin{equation}
\mathcal{L}_{per}^G= \|VGG_{L_9}(T_i)-VGG_{L_9}(E_i)\|_1
\end{equation}
where $VGG_{L_9}(P)$ means that the feature vector is taken from the ninth layer of a pre-trained VGG19 model for a given input $P$.\\
\textbf{Colorfulness Loss:} We observe that the combination of an $L_1$ loss and the perceptual loss cannot always produce satisfactory results and there is often a lack of colors in the colorized images. We identified that $L_1$ loss and perceptual loss often ignore the color consistency of the relatively smaller foreground objects. Minimization of $L_1$ alone often generates less saturated images. To overcome the problem, we propose a Kullback–Leibler (KL)-divergence-based loss function to measure the colorfulness of generated colorized images. We first convert $E^i$ and $T^i$ to their respective RGB color images. Next, we calculated the normalized histogram of each of the color channel for the respective images and define the colorfulness loss as   

\begin{equation}
\mathcal{L}_{colorfulness}^{G_m}= \sum_{i=1}^3 \|D_{kl}(hist_{T_{Ci}},hist_{E_{Ci}})\|
\end{equation}

where $hist_{X_{Ci}}$ means the normalized histogram of the $i^{th}$ channel of the image ($X$) and $D_{kl}$ is the KL-divergence.\\
We define the GAN loss $\mathcal{L}^{G_m}$ of the generator $G_m$ as
\begin{equation}
\mathcal{L}^{G_m}_{GAN}=\mathcal{L}_{BCE}(D_m(L^i,G(L^i,S^i)),1) 
\end{equation}

whereas the GAN loss for the discriminator $D_m$ is defined as
\begin{equation}
\begin{split}
\mathcal{L}^{D_m}_{GAN}=&\mathcal{L}_{BCE}(D_m(L^i,T^i),1)\\
 & +\mathcal{L}_{BCE}(D_m(L^i,G_m(L^i,S^i)),0)
\end{split}
\end{equation}

where $\mathcal{L}_{BCE}$ is the BCE loss function calculated as 

\begin{equation}
\mathcal{L}_{BCE} ={(y\log(p) + (1 - y)\log(1 - p))}
\end{equation}
where $y$ is the true label of a sample and $p$ is the predicted label of the sample.\\
In the IOC module, we add categorical cross-entropy as class-based object classification loss into the generator.
\begin{equation}
\mathcal{L}_{CE} =\sum_{i=1}^{N}{y_i.log (\hat{y}_i)}
\end{equation}
where $N$ is the total number of classes, $y_i$ and $\hat{y_i}$  are the actual class label of one hot ground truth and predicted label value, respectively.  The final loss $\mathcal{L}^{G_{ioc}}$ for the IOC module is calculated as
\begin{equation}
\mathcal{L}^{G_{ioc}}=\lambda_1{L}^G_1 + \lambda_2{L}_{per}^G + \lambda_3{L}_{colorfulness}^G +  \lambda_4{L}^G_{GAN}+\lambda_5{L}_{CE}
\end{equation}

where $\lambda_i$ is the weighting factor. In our work, we consider $\lambda_1=10$, $\lambda_2=1$, $\lambda_3=1$, $\lambda_4=1$ and $\lambda_5=1$.

Though discriminator $D_m$ can be trained directly by minimizing the loss function $\mathcal{L}^{D_m}_{GAN}$, we take a linear combination of the losses associated with the generator to calculate the fusion module generator loss $\mathcal{L}^{G_m}$ for training. The final loss $\mathcal{L}^{G_{m}}$ for the fusion module is calculated as
\begin{equation}
\mathcal{L}^{G_{m}}=\lambda_1{L}^G_1 + \lambda_2{L}_{per}^G + \lambda_3{L}_{colorfulness}^G + \lambda_4{L}^G_{GAN}
\end{equation}
where $\lambda_i$ is the weighting factor. In our work, we consider $\lambda_1=10$, $\lambda_2=1$, $\lambda_3=1$ and $\lambda_4=1$. \\
\section{Experimental Results}
\label{sec:results}
\subsection{Data set }
\begin{figure*}
  \centering
  \includegraphics[width=\linewidth]{ 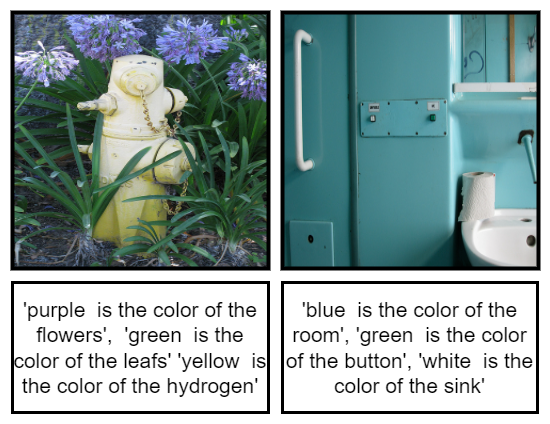}
  \caption{Examples of some samples of the  MS COCO dataset. }
    \label{fig:dataset}
\end{figure*}

\begin{figure*}
  \centering
  \includegraphics[width=\linewidth]{ 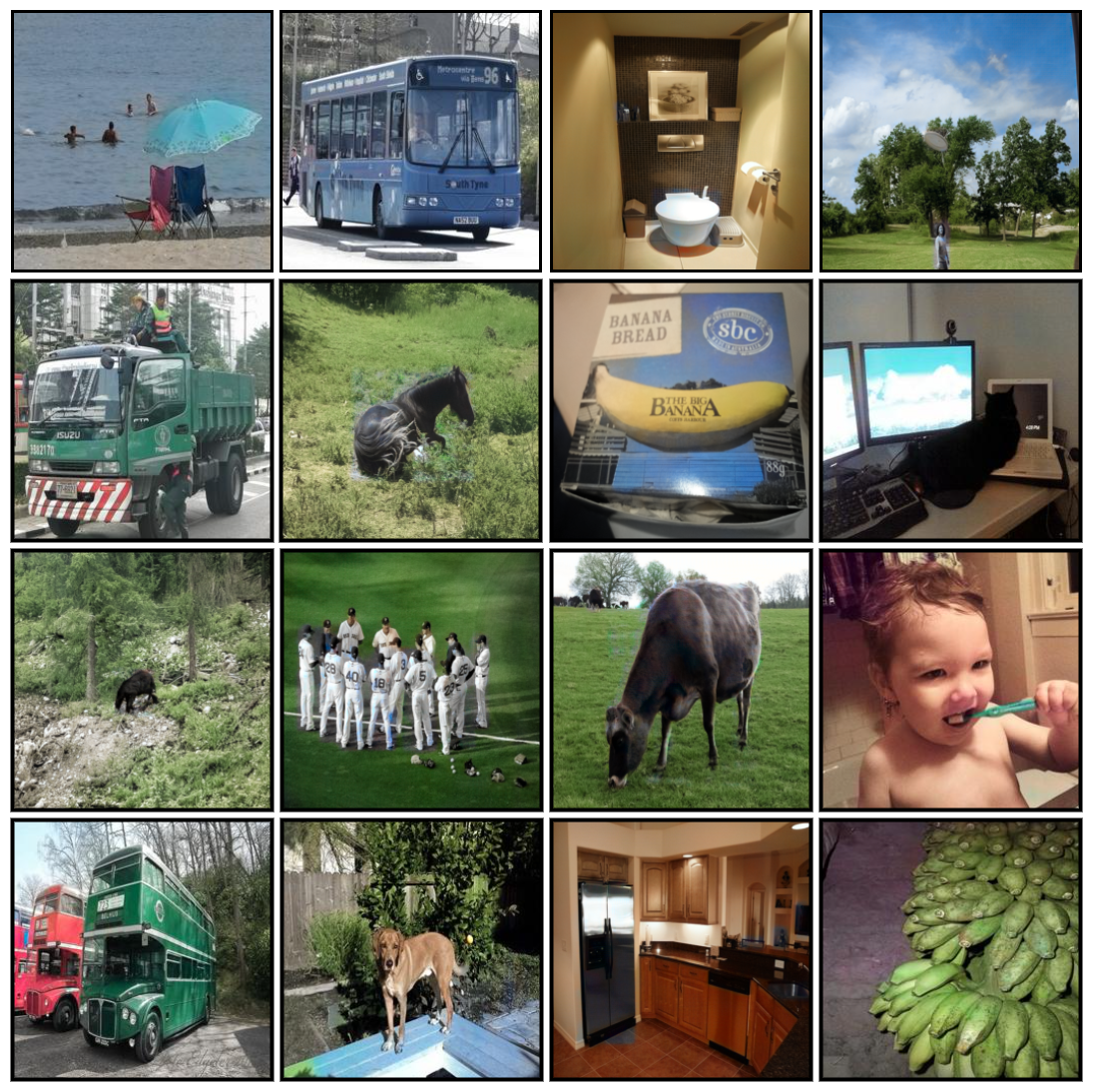}
  \caption{Examples of Some qualitative results generated by the proposed framework. [Best visible in $300\%$ zoom ]}
    \label{fig:result}
\end{figure*}
We use the MS-COCO-QA \cite{Ren2015ExploringMA} dataset for our training and evaluation. The MS COCO QA datatset contains 42429 images with color information. We use the color information as the auxiliary conditioning for the IOC module.  A total of 38136 images are used in training and 4293 images are used for testing. A few samples of the database are shown in Fig \ref{fig:dataset}.

\subsection{Qualitative results}

To evaluate the performance of our proposed method, we have generated a large number of images using the proposed framework. Some of the sample colorized results images are shown Fig \ref{fig:result}. It is observed that the proposed method is able to colorize complex scene images efficiently. As shown in Fig. \ref{fig:result}, the model can handle scenes with multiple objects, occlusions, shadows, etc. To further evaluate the quality of the colorized images, we have shown a set of 20 randomly selected images to 41 viewers where half of the images are natural RGB images and half of the images are colorized using the proposed approach. We asked the viewers to mark whether a displayed image is colorized or not within 5 seconds. In this experiment, we obtained an average accuracy of 53.41\%, which is just slightly better than random guessing. This uncontrolled experiment proves that the proposed method colorizes an image, maintaining the natural color consistency. 

\subsection{Comparison results}
\begin{figure*}
  \centering
  \includegraphics[width=\linewidth*.9]{ 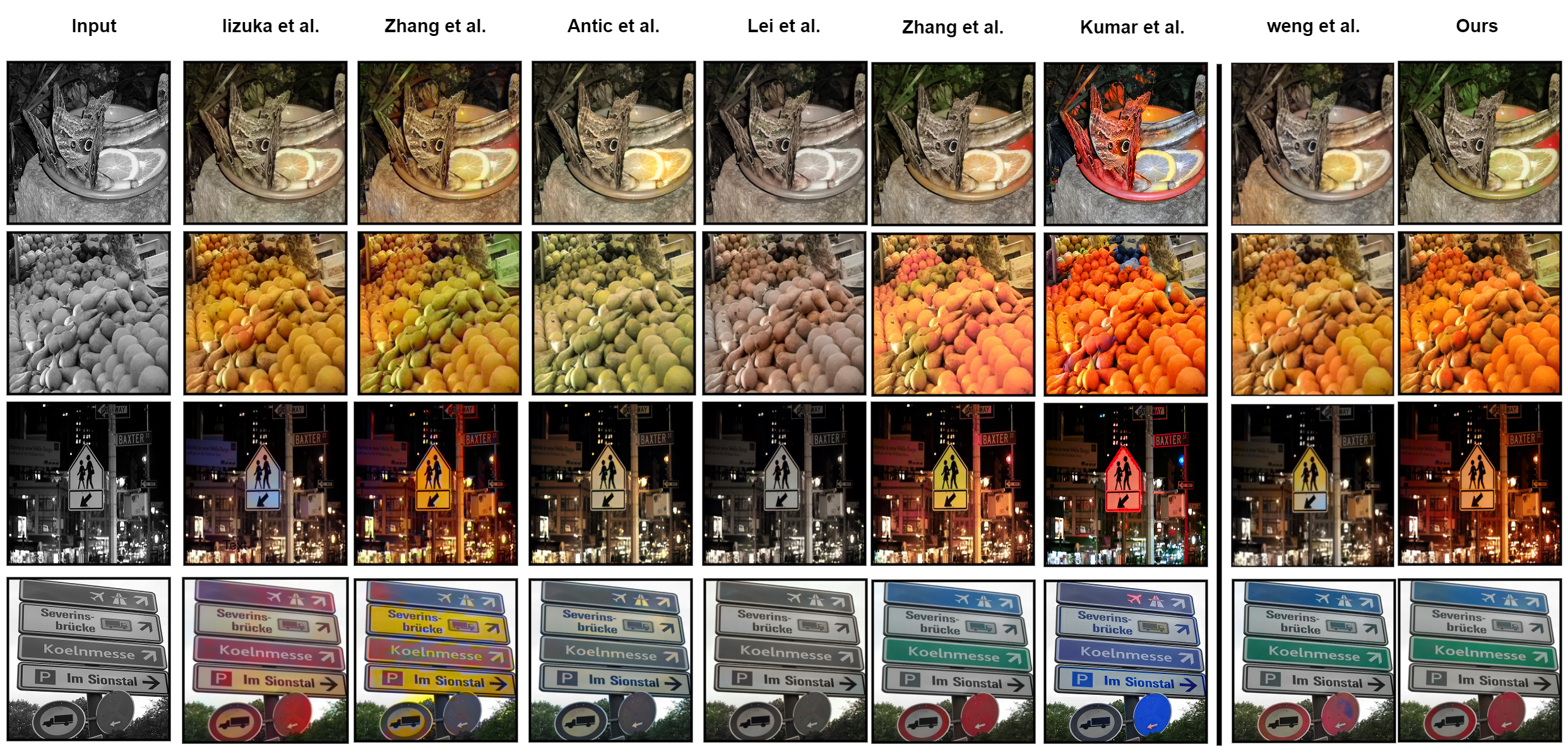}

  \caption{Qualitative comparison results: The first column contains input images, the  second column to eight column contain generated results of the SOTA algorithms, and the last column shows the results generated by the proposed algorithm. SOTA methods are  Iizuka et al.\cite{Iizuka2016LetTB} Zhang et al.\cite{Zhang2016ColorfulIC} Antic et al.\cite{deoldify} Lei et al.\cite{Lei2019FullyAV}   Zhang et al.\cite{Zhang2017RealtimeUI} Kumar et al.\cite{Kumar2021ColorizationT} weng et al.\cite{Weng2022LCoDeLC} . [Best visible in $300\%$ zoom ] }
    \label{fig:compare}
\end{figure*}

\begin{figure*}

  \centering
  \includegraphics[width=\linewidth ]{ 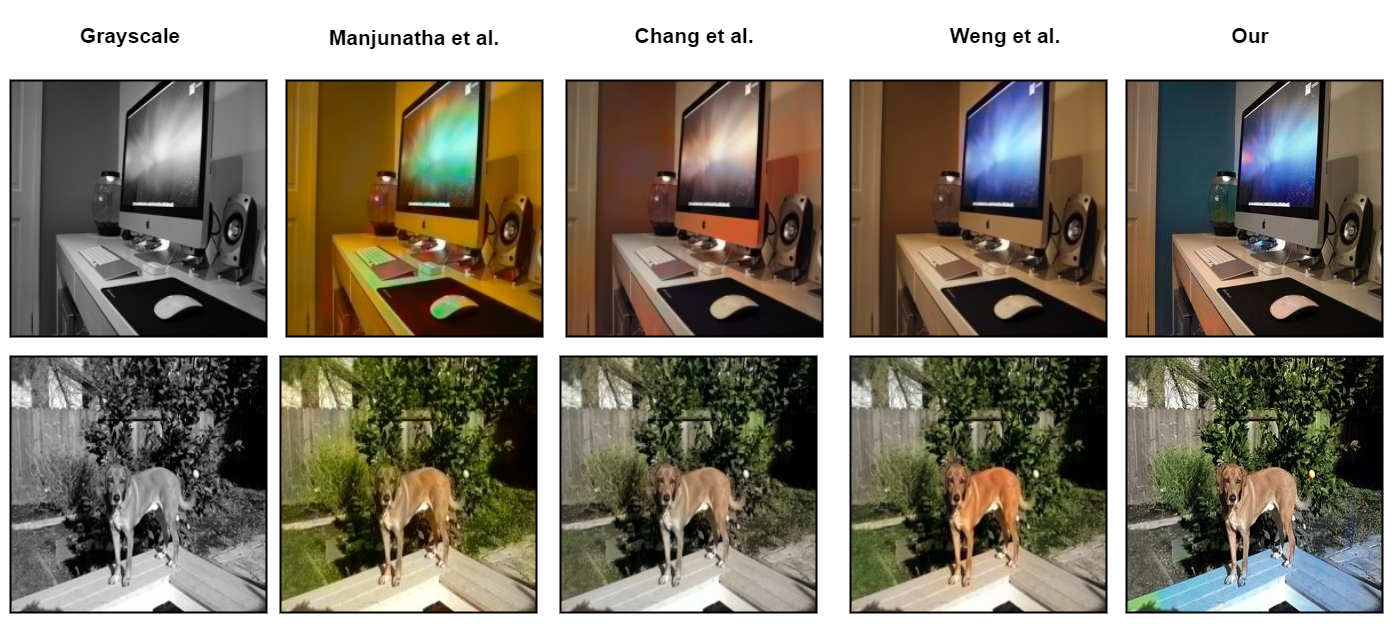}
 
  \caption{\textcolor{black}{Comparison results of the proposed algorithm with the existing  text-based colorization algorithms.}}
    \label{fig:com1}
\end{figure*}

To validate the effectiveness of the proposed method, we evaluate the  qualitative and quantitative results as well. In the Table \ref{tab:result}, we measured LPIPS\cite{simonyan2015very}, PSNR and SSIM \cite{wang2004image} to compare the proposed framework with the existing image colorization methods. We have considered the algorithms due to lizuka et al.\cite{Iizuka2016LetTB} Zhang et al.\cite{Zhang2016ColorfulIC} Antic et al.\cite{deoldify} Lei et al.\cite{Lei2019FullyAV}   Zhang et al.\cite{Zhang2017RealtimeUI} Kumar et al.\cite{Kumar2021ColorizationT} and weng et al.\cite{Weng2022LCoDeLC} for comparison. As shown in Table \ref{tab:result}, the proposed algorithm has outperformed the existing algorithms in terms of LPIPS, PSNR and SSIM scores. Furthermore, we also compared the qualitative results in Fig \ref{fig:compare}.\\
To further validate, we choose three text-based methods and used the same dataset in \cite{Weng2022LCoDeLC}. As shown in Table \ref{tab:result1}, the proposed algorithm has performed well in all the scores. We compare the qualitative results in Fig \ref{fig:com1}, and observe that except \cite{Weng2022LCoDeLC}, the colorized images, produced by \cite{Manjunatha2018LearningTC} and \cite{Chang2022LCoDerLC}, do not look natural. The images produced by \cite{Weng2022LCoDeLC} look over-saturated in color.

It can be seen from the Fig. \ref{fig:result} that the proposed algorithm generates perceptually consistent and well-colorized images. 

\begin{table}
\caption{Comparison results of the proposed algorithm with the existing colorization algorithms.}
\begin{tabular}{lllll}
\hline
                & LPIPS $\downarrow$ & PSNR $\uparrow$   & SSIM $\uparrow$  &  \\ \cline{1-4}
lizuka et al.\cite{Iizuka2016LetTB}   & 0.185 & 23.860  & 0.922 &  \\ 
Zhang et al.\cite{Zhang2016ColorfulIC}    & 0.234 & 21.838 & 0.895  &  \\ 
Antic et al.\cite{deoldify} & 0.180 & 23.692 & 0.920 &  \\ 
Lei et al.\cite{Lei2019FullyAV}      & 0.191 & 24.588 & 0.922 &  \\ 
Zhang et al.\cite{Zhang2017RealtimeUI}    & 0.138 & 26.823 & 0.937 &  \\ 
Kumar et 
 al.\cite{Kumar2021ColorizationT}       & 0.137 & 26.653 & 0.937 & \\ \cline{1-4}
 Weng et al.\cite{Weng2022LCoDeLC}       & 0.138 & 27.329 & 0.921 & \\
Ours            & \textbf{0.120} & \textbf{28.214} & \textbf{0.938} &  \\ 
\hline
\end{tabular}

\label{tab:result}
\end{table}
\begin{figure*}
  \centering
  \includegraphics[width=\linewidth]{ 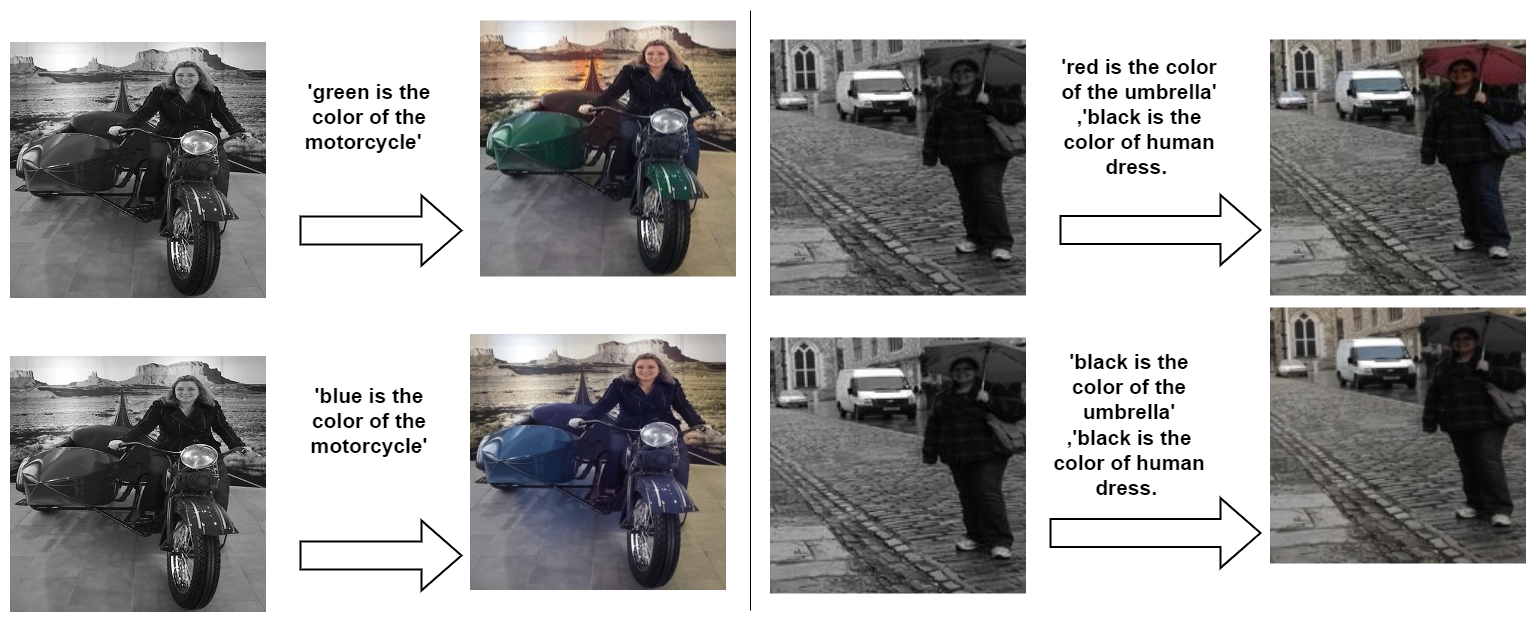}
  \caption{Examples of object re-colorization. [Best visible in $300\%$ zoom ]}
    \label{fig:text_ablation}
\end{figure*}

\begin{table}
\caption{\textcolor{black}{Comparison results of the proposed algorithm with the existing  text-based colorization algorithms. The dataset split of Weng et al.\cite{Weng2022LCoDeLC}} is used to generate the results.}
\small
\begin{tabular}{lllll}
\hline
 & Method  & LPIPS $\downarrow$ & PSNR $\uparrow$   & SSIM $\uparrow$  \\ \cline{1-5}
 & Manjunatha et. al \cite{Manjunatha2018LearningTC}& 0.282 & 21.055 & 0.853 \\
 & Chang et al.\cite{Chang2022LCoDerLC} & 0.159 & 25.504 & 0.917 \\
 & Weng et al.\cite{Weng2022LCoDeLC} & 0.169 & 24.965 & 0.917 \\ \cline{1-5}
 & Ours    & \textbf{0.128} & \textbf{27.230} & \textbf{0.933}
\end{tabular}
\label{tab:result1}
\end{table}

\subsection{Ablation study}
\begin{figure}
  \centering
  \includegraphics[width=\linewidth ]{ 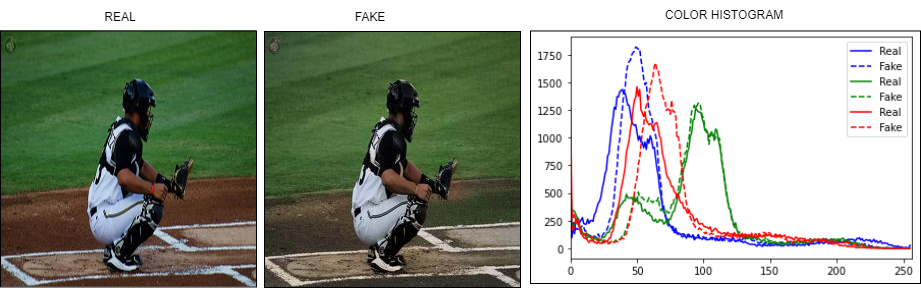}
  \caption{\textcolor{black}{Channel wise histogram. [Best visible in $300\%$ zoom ]}}
    \label{fig:histo}
\end{figure}

To further validate the contributions of the novel components that are present in our frameworks, we perform a detailed ablation study. We mainly focus on the effectiveness of textual conditioning along with the effect of the proposed colorfulness loss term. As shown in Table \ref{tab:ablation}, we observe the presence of textual conditioning significantly improves all three metrics. Though the proposed Colorfulness (CF) Loss slightly improves the LPIPS score, the improvements in PSNR and SSIM scores are significantly high when the colorfulness loss is incorporated.  Our ablation studies showed that we achieved the best result when both colorfulness loss and textual information were used in the colorization framework. Finally, to validate the usefulness of the proposed algorithm on the overall generation quality, we observe the alignment of the color histograms of the real GT image and the colorized image. As shown in Fig. \ref{fig:histo}, the color histogram of the colorized image closely follows the color histogram of the ground truth. 

\begin{table}
\caption{Ablation results of different components used in our framework. \textit{Text Info} indicates the usage of auxiliary text information and \textit{CF Loss} indicates the proposed colorfulness loss.}
\begin{tabular}{lcclll}
\hline 
 Text Info & CF Loss & LPIPS $\downarrow$ & PSNR $\uparrow$   & SSIM $\uparrow$  \\ \cline{1-5}
 X                 & X              & 0.156 & 24.230 & 0.896 \\
 X                 &  $ \checkmark $  & 0.155 & 24.625 & 0.912 \\
$ \checkmark $      & X              & 0.130 & 27.238 & 0.921 \\
$ \checkmark $      & $ \checkmark $  & 0.120 & 28.214 & 0.938 \\\cline{1-5}
 
\end{tabular}
\label{tab:ablation}
\end{table}

\subsection{Re-Colorization}
As we use auxiliary conditioning for the colorization process, we can modulate the textual information to re-colorize different object instances of the same input image. Some re-colorization results are shown in Fig.
\ref{fig:text_ablation}. As shown in the figure, the proposed methods can seamlessly handle re-colorization task depending on the textual conditioning.

\subsection{Failure Cases}
\begin{figure}
  \centering
  \includegraphics[width=0.8\linewidth]{ 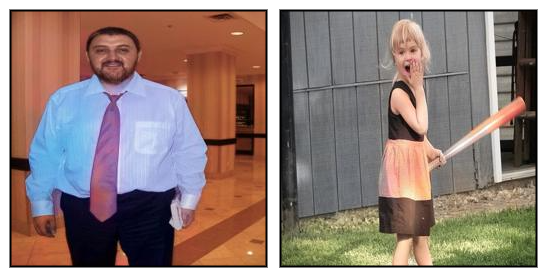}
  \caption{Examples of some failure cases of the proposed method.}
    \label{fig:fail}
\end{figure}

As shown in Fig. \ref{fig:result}, the proposed algorithm works considerably well even for complex scenes containing multiple objects. However, in certain cases, the proposed framework fails to generate satisfactory results. Upon investigation, we found that this often happens when the sample does not have enough auxiliary textual descriptions of the colors. Some failure cases are shown in Fig. \ref{fig:fail}. This problem, however, can be mitigated by introducing more color descriptions for the object instances.

\section{Conclusion}
\label{sec:conclusion}
In this paper, a novel  image colorization approach is proposed that utilises the color information as an auxiliary conditioning of the network.  The results of the proposed algorithm are compared with the existing image colorization methods, and it can be seen that the proposed method outperforms the existing ones in terms of LPIPS and PSNR metrics. We validated that the object instance-level colorization produces superior results when auxiliary textual conditioning is available. A novel loss function has also been introduced to have more fidelity in the color generation process. In certain cases, it is observed that the proposed method  produces less colorful images when the color information is not adequately present in the conditioning texts. This problem can be solved in future by adding additional text description in the database.

{\small
\bibliographystyle{ieee_fullname}
\bibliography{egbib}

\begin{thebibliography}{10}\itemsep=-1pt

\bibitem{deoldify}
Jason Antic.
\newblock A deep learning based project for colorizing and restoring old images
  (and video!). https://github.com/jantic/deoldify,, 2019.

\bibitem{anwar2020ColorSurvey}
Saeed Anwar, Muhammad Tahir, Chongyi Li, Ajmal Mian, Fahad~Shahbaz Khan, and
  Abdul~Wahab Muzaffar.
\newblock Image colorization: A survey and dataset.
\newblock {\em arXiv preprint arXiv:2008.10774}, 2020.

\bibitem{Bahng2018ColoringWW}
Hyojin Bahng, Seungjoo Yoo, Wonwoong Cho, David~Keetae Park, Ziming Wu,
  Xiaojuan Ma, and Jaegul Choo.
\newblock Coloring with words: Guiding image colorization through text-based
  palette generation.
\newblock In {\em ECCV}, 2018.

\bibitem{Bastos2013RUNTIMEGS}
Rui Bastos, William~C. Wynn, and Anselmo Lastra.
\newblock Run-time glossy surface self-transfer processing.
\newblock In {\em MULTIMEDIA}, 2013.

\bibitem{Caesar2018COCOStuffTA}
Holger Caesar, Jasper R.~R. Uijlings, and Vittorio Ferrari.
\newblock Coco-stuff: Thing and stuff classes in context.
\newblock {\em 2018 IEEE/CVF Conference on Computer Vision and Pattern
  Recognition}, pages 1209--1218, 2018.

\bibitem{Carlucci2018DE2CODD}
Fabio~Maria Carlucci, Paolo Russo, and Barbara Caputo.
\newblock $(de)^2co$: Deep depth colorization.
\newblock {\em IEEE Robotics and Automation Letters}, 2018.

\bibitem{Chang2022LCoDerLC}
Zheng Chang, Shuchen Weng, Yu Li, Si Li, and Boxin Shi.
\newblock L-coder: Language-based colorization with color-object decoupling
  transformer.
\newblock In {\em ECCV}, 2022.

\bibitem{Cheng2015DeepC}
Zezhou Cheng, Qingxiong Yang, and Bin Sheng.
\newblock Deep colorization.
\newblock {\em 2015 IEEE International Conference on Computer Vision (ICCV)},
  pages 415--423, 2015.

\bibitem{Deng2009ImageNetAL}
Jia Deng, Wei Dong, Richard Socher, Li-Jia Li, K. Li, and Li Fei-Fei.
\newblock Imagenet: A large-scale hierarchical image database.
\newblock {\em 2009 IEEE Conference on Computer Vision and Pattern
  Recognition}, pages 248--255, 2009.

\bibitem{Devlin2019BERTPO}
Jacob Devlin, Ming-Wei Chang, Kenton Lee, and Kristina Toutanova.
\newblock Bert: Pre-training of deep bidirectional transformers for language
  understanding.
\newblock {\em ArXiv}, abs/1810.04805, 2019.

\bibitem{Girdhar2018DetectandTrackEP}
Rohit Girdhar, Georgia Gkioxari, Lorenzo Torresani, Manohar Paluri, and Du
  Tran.
\newblock Detect-and-track: Efficient pose estimation in videos.
\newblock {\em 2018 IEEE/CVF Conference on Computer Vision and Pattern
  Recognition}, pages 350--359, 2018.

\bibitem{Huang2022DeepLF}
Shanshan Huang, Xin Jin, Qian Jiang, and Li Liu.
\newblock Deep learning for image colorization: Current and future prospects.
\newblock {\em Eng. Appl. Artif. Intell.}, 114:105006, 2022.

\bibitem{Huang2005AnAE}
Yi-Chin Huang, Yi-Shin Tung, Jun-Cheng Chen, Sung-Wen Wang, and Ja-Ling Wu.
\newblock An adaptive edge detection based colorization algorithm and its
  applications.
\newblock In {\em MULTIMEDIA '05}, 2005.

\bibitem{Iizuka2016LetTB}
Satoshi Iizuka, Edgar Simo-Serra, and Hiroshi Ishikawa.
\newblock Let there be color!
\newblock {\em ACM Transactions on Graphics (TOG)}, 35:1 -- 11, 2016.

\bibitem{Kumar2021ColorizationT}
Manoj Kumar, Dirk Weissenborn, and Nal Kalchbrenner.
\newblock Colorization transformer.
\newblock {\em ArXiv}, abs/2102.04432, 2021.

\bibitem{Lei2019FullyAV}
Chenyang Lei and Qifeng Chen.
\newblock Fully automatic video colorization with self-regularization and
  diversity.
\newblock {\em 2019 IEEE/CVF Conference on Computer Vision and Pattern
  Recognition (CVPR)}, pages 3748--3756, 2019.

\bibitem{Levin2004ColorizationUO}
Anat Levin, Dani Lischinski, and Yair Weiss.
\newblock Colorization using optimization.
\newblock In {\em SIGGRAPH 2004}, 2004.

\bibitem{Luo2022ThermalII}
Fuya Luo, Yunhan Li, Guang Zeng, Peng Peng, Gang Wang, and Yongjie Li.
\newblock Thermal infrared image colorization for nighttime driving scenes with
  top-down guided attention.
\newblock {\em IEEE Transactions on Intelligent Transportation Systems},
  23:15808--15823, 2022.

\bibitem{Maas2013RectifierNI}
Andrew~L Maas, Awni~Y Hannun, Andrew~Y Ng, et~al.
\newblock Rectifier nonlinearities improve neural network acoustic models.
\newblock In {\em Proc. icml}, page~3. Atlanta, Georgia, USA, 2013.

\bibitem{Manjunatha2018LearningTC}
Varun Manjunatha, Mohit Iyyer, Jordan~L. Boyd-Graber, and Larry~S. Davis.
\newblock Learning to color from language.
\newblock In {\em NAACL}, 2018.

\bibitem{Perazzi2016ABD}
Federico Perazzi, Jordi Pont-Tuset, Brian McWilliams, Luc~Van Gool, Markus~H.
  Gross, and Alexander Sorkine-Hornung.
\newblock A benchmark dataset and evaluation methodology for video object
  segmentation.
\newblock {\em 2016 IEEE Conference on Computer Vision and Pattern Recognition
  (CVPR)}, pages 724--732, 2016.

\bibitem{Ren2015ExploringMA}
Mengye Ren, Ryan Kiros, and Richard~S. Zemel.
\newblock Exploring models and data for image question answering.
\newblock In {\em NIPS}, 2015.

\bibitem{simonyan2015very}
Karen Simonyan and Andrew Zisserman.
\newblock Very deep convolutional networks for large-scale image recognition.
\newblock In {\em The International Conference on Learning Representations
  (ICLR)}, 2015.

\bibitem{Su2020InstanceAwareIC}
Jheng-Wei Su, Hung kuo Chu, and Jia-Bin Huang.
\newblock Instance-aware image colorization.
\newblock {\em 2020 IEEE/CVF Conference on Computer Vision and Pattern
  Recognition (CVPR)}, pages 7965--7974, 2020.

\bibitem{Tola2008AFL}
Engin Tola, Vincent Lepetit, and Pascal~V. Fua.
\newblock A fast local descriptor for dense matching.
\newblock {\em 2008 IEEE Conference on Computer Vision and Pattern
  Recognition}, pages 1--8, 2008.

\bibitem{Treneska2022GANBasedIC}
Sandra Treneska, Eftim Zdravevski, I. Pires, Petre Lameski, and Sonja Gievska.
\newblock Gan-based image colorization for self-supervised visual feature
  learning.
\newblock {\em Sensors (Basel, Switzerland)}, 22, 2022.

\bibitem{Wang2018GeneratingHQ}
Puyang Wang and Vishal~M. Patel.
\newblock Generating high quality visible images from sar images using cnns.
\newblock {\em 2018 IEEE Radar Conference (RadarConf18)}, pages 0570--0575,
  2018.

\bibitem{wang2004image}
Zhou Wang, Alan~C Bovik, Hamid~R Sheikh, and Eero~P Simoncelli.
\newblock Image quality assessment: From error visibility to structural
  similarity.
\newblock {\em IEEE Transactions on Image Processing (TIP)}, 2004.

\bibitem{WelinderEtal2010}
P. Welinder, S. Branson, T. Mita, C. Wah, F. Schroff, S. Belongie, and P.
  Perona.
\newblock {Caltech-UCSD Birds 200}.
\newblock Technical Report CNS-TR-2010-001, California Institute of Technology,
  2010.

\bibitem{Weng2022LCoDeLC}
Shuchen Weng, Hao Wu, Zheng Chang, Jiajun Tang, Si Li, and Boxin Shi.
\newblock L-code: Language-based colorization using color-object decoupled
  conditions.
\newblock In {\em AAAI}, 2022.

\bibitem{Wu2022FinegrainedSE}
Dirbaba Niguse Bekele~Hongjuan Wu, Jianhou Gan, Juxiang Zhou, Jun Wang, and Wei
  Gao.
\newblock Fine‐grained semantic ethnic costume high‐resolution image
  colorization with conditional gan.
\newblock {\em International Journal of Intelligent Systems}, 37:2952 -- 2968,
  2022.

\bibitem{Wu2021TowardsVA}
Yanze Wu, Xintao Wang, Yu Li, Honglun Zhang, Xun Zhao, and Ying Shan.
\newblock Towards vivid and diverse image colorization with generative color
  prior.
\newblock {\em 2021 IEEE/CVF International Conference on Computer Vision
  (ICCV)}, pages 14357--14366, 2021.

\bibitem{Xiao2022SemanticawareAI}
Yuxuan Xiao, Aiwen Jiang, Changhong Liu, and Mingwen Wang.
\newblock Semantic‐aware automatic image colorization via unpaired
  cycle‐consistent self‐supervised network.
\newblock {\em International Journal of Intelligent Systems}, 37:1222 -- 1238,
  2022.

\bibitem{Zhang2016ColorfulIC}
Richard Zhang, Phillip Isola, and Alexei~A. Efros.
\newblock Colorful image colorization.
\newblock In {\em ECCV}, 2016.

\bibitem{Zhang2017RealtimeUI}
Richard Zhang, Jun-Yan Zhu, Phillip Isola, Xinyang Geng, Angela~S. Lin, Tianhe
  Yu, and Alexei~A. Efros.
\newblock Real-time user-guided image colorization with learned deep priors.
\newblock {\em ACM Transactions on Graphics (TOG)}, 36:1 -- 11, 2017.

\end{thebibliography}
}

\end{document}